\documentclass{article} 
\usepackage{iclr2019_conference,times}

\usepackage[pdftex]{graphicx}
\usepackage{booktabs}
\usepackage{amsmath}
\usepackage{bm}
\usepackage{amssymb}
\usepackage{multirow}
\usepackage{caption}
\usepackage{appendix}
\usepackage{pifont}
\usepackage{mathtools}
\usepackage{todonotes}
\usepackage{bbm}
\usepackage{subcaption}

\usepackage{hyperref}
\usepackage{url}

\title{Just Jump: Dynamic Neighborhood\\Aggregation in Graph Neural Networks}

\author{Matthias Fey\\
Department of Computer Graphics\\
TU Dortmund University\\
44227 Dortmund, Germany\\
\texttt{matthias.fey@udo.edu}
}

\def\eg{\emph{e.g.}}   
\def\ie{\emph{i.e.}}   
\def\cf{\emph{cf.}}     

\newcommand{\lp}{\hspace{-2pt}\left(}
\newcommand{\rp}{\right)\hspace{-2pt}}
\newcommand{\lbr}{\hspace{-2pt}\left[}
\newcommand{\rbr}{\right]\hspace{-2pt}}
\newcommand{\lmid}{\hspace{-2pt}\left|}
\newcommand{\rmid}{\right|\hspace{-2pt}}
\newcommand{\concat}{\,\Vert\,}

\newcommand{\mr}[2]{\multirow{#1}{*}{#2}}
\newcommand{\rb}[2]{\rotatebox{#1}{#2}}

\newcommand{\bo}[1]{\textbf{#1}}
\newcommand{\gr}[1]{\textcolor{black!70!white}{#1}}
\newcommand{\no}[1]{#1}

\iclrfinalcopy{}  
\begin{document}

\maketitle

\begin{abstract}
  We propose a \emph{dynamic neighborhood aggregation} (DNA) procedure guided by (multi-head) attention for representation learning on graphs.
  In contrast to current graph neural networks which follow a simple neighborhood aggregation scheme, our DNA procedure allows for a selective and node-adaptive aggregation of neighboring embeddings of potentially differing locality.
  In order to avoid overfitting, we propose to control the channel-wise connections between input and output by making use of grouped linear projections.
  In a number of transductive node-classification experiments, we demonstrate the effectiveness of our approach.
\end{abstract}

\section{Introduction and Related Work}%
\label{sec:introduction_and_related_work}

Graph neural networks (GNNs) have become the de facto standard for representation learning on relational data \citep{Bronstein/etal/2017,Gilmer/etal/2017,Battaglia/etal/2018}.
GNNs follow a simple neighborhood aggregation procedure motivated by two major perspectives: The generalization of classical CNNs to irregular domains \citep{Shuman/etal/2013}, and their strong relations to the \citet{Weisfeiler/Lehman/1968} algorithm \citep{Xu/etal/2019,Morris/etal/2019}.
Many different graph neural network variants have been proposed and significantly advanced the state-of-the-art in this field \citep{Defferrard/etal/2016,Kipf/Welling/2017,Monti/etal/2017,Gilmer/etal/2017,Hamilton/etal/2017,Velickovic/etal/2018,Fey/etal/2018}.

Most of these approaches focus on novel kernel formulations, however, deeply stacking those layers usually result in gradually decreasing performance despite having, in principal, access to a wider range of information \citep{Kipf/Welling/2017}.
\citet{Xu/etal/2018} blame the strongly varying speed of expansion on this phenomenon, caused by locally differing graph structures, and hence propose to node-adaptively \emph{jump} back to earlier representations if those fit the task at hand more precisely.

Inspired by these so-called Jumping Knowledge networks \citep{Xu/etal/2018}, we explore a highly dynamic neighborhood aggregation (DNA) procedure based on scaled dot-product attention \citep{Vaswani/etal/2017} which is able to aggregate neighboring node representations of differing locality.
We show that this approach, when additionaly combined with grouped linear projections, outperforms traditional stacking of GNN layers, even when those are enhanced by Jumping Knowledge.

We briefly give a formal overview of the related work before we propose our method in more detail:

\paragraph{Graph Neural Networks (GNNs)}%
\label{par:graph-neural-networks}

operate over graph structured data $\mathcal{G} = (\mathcal{V}, \mathcal{E})$ and iteratively update node features $\vec{h}_v^{(t-1)}$ of node $v \in \mathcal{V}$ in layer $t-1$ by aggregating localized information via
\begin{equation}
  \vec{h}_v^{(t)} = f_{\bm{\Theta}}^{(t)} \lp h_v^{(t-1)}, {\left\{  h_w^{(t-1)} \right\}}_{w \in \mathcal{N}(v)} \rp, \quad \textrm{\eg,} \quad \vec{h}_v^{(t)} = \sigma \lp \bm{\Theta}^{(t)} \sum_{\mathclap{w \in \mathcal{N}(v) \cup \{ v \}}} C_{v,w}^{(t-1)} \, \vec{h}_w^{(t-1)} \rp,
\end{equation}
from the neighbor set $\mathcal{N}(\cdot)$ through a differentiable function $f_{\bm{\Theta}}^{(t)}$ parametrized by weights ${\bm{\Theta}}^{(t)}$.
In current implementations, $C_{v,w}^{(t-1)}$ is either defined to be static \citep{Xu/etal/2019}, structure- \citep{Kipf/Welling/2017,Hamilton/etal/2017} or data-dependent \citep{Velickovic/etal/2018}.

GNN layers are typically stacked sequentially, but can be optionally enhanced by skip connections, \eg, $\vec{h}_v^{(t)} \leftarrow \vec{h}_v^{(t)} + \bm{\Theta}_s^{(t)} \vec{h}_v^{(t-1)}$ \citep{Cangea/etal/2018}, or updated using Gated Recurrent Units via $\vec{h}_v^{(t)} \leftarrow \textrm{GRU}(\vec{h}_v^{(t-1)}, \vec{h}_v^{(t)})$ \citep{Cho/etal/2014,Li/etal/2016}.
After $T$ layers, $\vec{h}_v^{(T)}$ holds the $T$-hop subgraph representation centered around node $v$.

\paragraph{Jumping Knowledge (JK) networks}%
\label{par:jumping_knowledge_networks}

enable deeper GNNs by introducing layer-wise jump connections and selective aggregations to leverage node-adaptive neighborhood ranges~\citep{Xu/etal/2018}.
Given layer-wise representations $\vec{h}_v^{(1)}, \ldots, \vec{h}_v^{(T)}$ of node $v$, its final output representation is obtained by either
\vspace{-3pt}
\begin{equation}
  \arraycolsep=10pt
  \def\arraystretch{1.7}
  \begin{array}{ccccc}
    \textrm{(1)}~\textbf{concatenation}, & & \textrm{(2)}~\textbf{pooling} & \textrm{or} & \textrm{(3)}~\textbf{summation}\\
    \vec{h}_v^{(1)} \concat \ldots \concat \vec{h}_v^{(T)} & & \max \lp \vec{h}_v^{(1)}, \ldots, \vec{h}_v^{(T)} \rp & & \sum\nolimits_{t=1}^T \alpha_v^{(t)} \vec{h}_v^{(t)}
  \end{array}
\end{equation}
where scorings $\alpha_v^{(t)}$ are obtained from a bi-directional LSTM~\citep{Hochreiter/Schmidhuber/1997}.

\paragraph{Attention modules}%
\label{par:attention-modules}

weight the values of a set of key-value pairs $\bm{K}, \bm{V} \in \mathbb{R}^{n \times d}$ according to a given query $\vec{q} \in \mathbb{R}^d$ by computing scaled dot-products between key-query pairs and using the softmax-normalized results as weighting coefficients \citep{Vaswani/etal/2017}:
\begin{equation}
  \textrm{Attention}(\vec{q}, \bm{K}, \bm{V}) = \textrm{softmax} \lp \frac{\vec{q}^{\top} \bm{K}^{\top}}{\sqrt{d}} \rp \bm{V}
\end{equation}
In practice, the attention function is usually performed $h$ times (with each head learning separate attention weights and attending to different positions) and the results are concatenated.

\paragraph{Grouped operations}%
\label{par:grouped-operations}

control the channel-wise connections between an input $\bm{X} \in \mathbb{R}^{n \times c}$ and an output $\bm{Y} \in \mathbb{R}^{n \times d}$ to reduce the number of parameters by $g$, the number of groups \citep{Krizhevsky/etal/2012}.
If $c = g$, the operation is performed independently over every channel \citep{Chollet/2017}.

\section{Method}%
\label{sec:method}

Closely related to the JK networks \citep{Xu/etal/2018}, we are seeking for a way to node-adaptively craft receptive-fields for a specific task at hand.
JK nets achieve this by dynamically jumping to the most representive layer-wise embedding \emph{after} a fixed range of node representations were obtained.
Hence, Jumping Knowledge can not guarantee that higher-order features will not become ``washed out'' in later layers, but instead will just fall back to more localized information preserved from earlier representations.
In addition, fine-grained details may still get lost very early on in expander-like subgraph structures \citep{Xu/etal/2018}.

In contrast, we propose to allow jumps to earlier knowledge immediately \emph{while} aggregating information from neighboring nodes.
This results in a highly-dynamic receptive-field in which neighborhood information is potentially gathered from representations of differing locality.
Each node's representation controls its own spread-out, possibly aggregating more global information in one branch, and falling back to more local information in others.

\begin{figure}[t]
  \centering
  \vspace{-1.0cm}
  \includegraphics{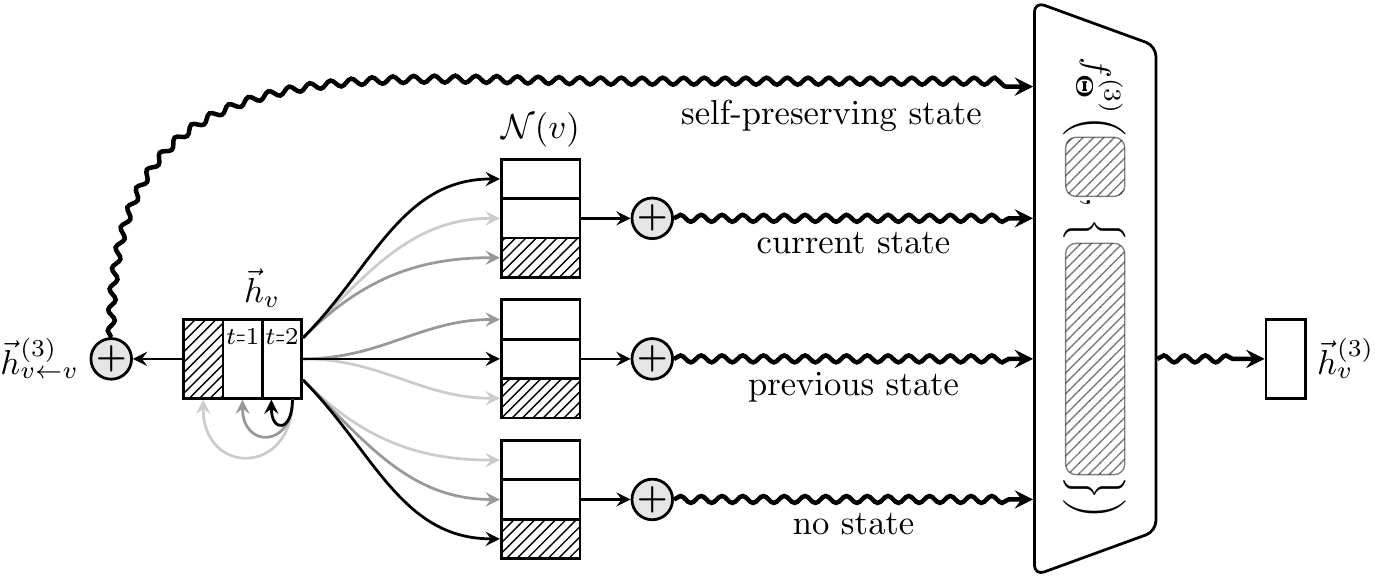}
  \caption{
    Given current node representation $\vec{h}_v^{(2)}$ as query, a node-adaptive embedding $\vec{h}_{v \leftarrow w}^{(3)}$ gets computed for all neighbors $w \in \mathcal{N}(v)$ based on their former representations $\vec{h}_{w}^{(1)}$ and $\vec{h}_w^{(2)}$, either preserving current state, previous state, or no state at all.
    In addition, self-attention is applied to retain central node information.
  }\label{fig:overview}
\end{figure}

Formally, we allow each node-neighborhood pair $(v,w) \in \mathcal{E}$ to attend to \emph{all} its former representations $\vec{h}^{(1)}_w, \ldots, \vec{h}^{(t-1)}_w$ while using its output $\vec{h}^{(t)}_{v \leftarrow w}$ for aggregation:
\begin{equation}
\begin{aligned}
  \vec{h}_v^{(t)} &= f_{\bm{\Theta}}^{(t)} \lp \vec{h}_{v \leftarrow v}^{(t)}, {\left\{ \vec{h}_{v \leftarrow w}^{(t)} \right\}}_{w \in \mathcal{N}(v)} \rp \textrm{~where}\\
  \vec{h}_{v \leftarrow w}^{(t)} &= \textrm{Attention} \lp {\lp \vec{h}_v^{(t-1)}\rp}^{\top} \bm{\Theta}_{Q}^{(t)}, \hspace{5pt} {\lbr \vec{h}_w^{(1)}, \ldots, \vec{h}_w^{(t-1)} \rbr}^{\top} \bm{\Theta}_{K}^{(t)}, \hspace{5pt} {\lbr \vec{h}_w^{(1)}, \ldots, \vec{h}_w^{(t-1)} \rbr}^{\top} \bm{\Theta}_{V}^{(t)} \rp
\end{aligned}
\end{equation}
with $\bm{\Theta}_Q^{(t)}, \bm{\Theta}_K^{(t)}, \bm{\Theta}_V^{(t)} \in \mathbb{R}^{d \times d}$ denoting trainable symmetric projection matrices.
A scheme of this layer is depicted in Figure~\ref{fig:overview}.
By ensuring that former information is preserved, our operator can be stacked deep by design, in particular without the need of JK nets.

In practice, we replace the single attention module by multi-head attention with a user-defined number of heads $h$ while maintaining the same number of parameters.
We implemented $f_{\bm{\Theta}}^{(t)}$ as the graph convolutional operator from \citet{Kipf/Welling/2017}, although any other GNN layer may be applicable.
Due to being already projected, we do not transform incoming node embeddings in $f_{\bm{\Theta}}^{(t)}$.

Furthermore, we incorporate an additional parameter to the softmax distribution of the attention module to allow the model to refuse the aggregation of individual neighboring embeddings in order to preserve fine-grained details (\cf~Figure~\ref{fig:overview}).
Instead of actually overparametrizing the resulting distribution, we restrict this parameter to be fixed \citep{Goodfellow/etal/2016}. 
This results in a softmax function of the form
\begin{equation}
  {\textrm{softmax} \lp \vec{x} \rp}_{\hspace{2pt} i} = \frac{\exp(x_i)}{1 + \sum_j \exp(x_j)} \in \mathbb{R}^n, \quad \textrm{where } \vec{x} \in \mathbb{R}^n.
\end{equation}

\paragraph{Feature dimensionality.}%
\label{par:feature_dimensionality.}

In order to leverage the attention module, input and output feature dimensionality are forced to remain equal across \emph{all} layers.
We found this to be only a weak constraint since this is already common practice \citep{Gilmer/etal/2017, Xu/etal/2018}.

\paragraph{Regularization.}%
\label{par:regularization}

We apply dropout \citep{Srivastava/etal/2014} to the softmax-normalized attention weights and use grouped linear projections with $g$ groups to reduce the number of parameters from $d^2$ to $d^2 \hspace{-1pt}/g$, where $g$ must be chosen so that $\max(g, h)$ is divisible by $\min(g, h)$.
The grouped projections regulate the attention heads by forcing them to only have a local influence on other attention heads (or even restricting them to have no influence at all).
We observed that these adjustments greatly help the model to avoid overfitting while still maintaining large effective hidden sizes.

\paragraph{Runtime.}%
\label{par:Runtime}

Our proposed operator does scale linearly in the number of previously seen node representations for each edge, \ie~$\mathcal{O}(|\mathcal{E}| T)$.
To account for large $T$, we suggest to restrict the inputs of the attention module to a fixed-sized subset of former representations.

\section{Experiments}%
\label{sec:experiments}

We evaluate our approach on 8 transductive benchmark datasets: the tasks of classifying academic papers (Cora, CiteSeer, PubMed, Cora Full) \citep{Sen/etal/2008, Bojchevski/Guennemann/2018}, active research fields of authors (Coauthor CS, Coauthor Physics) \citep{Shchur/etal/2018} and product categories (Amazon Computers, Amazon Photo) \citep{Shchur/etal/2018}.
We randomly split nodes into $20\%$, $20\%$ and $60\%$ for training, validation and testing.
Descriptions and statistics of all datasets can be found in Appendix~\ref{sec:datasets}.
The code with all its evaluation examples is integrated into the \emph{PyTorch Geometric}\footnote{\url{https://github.com/rusty1s/pytorch_geometric}} library \citep{Fey/Lenssen/2019}.

\paragraph{Setup.}%
\label{par:setup.}

We compare our DNA approach to GCN \citep{Kipf/Welling/2017} and GAT \citep{Velickovic/etal/2018} with and without Jumping Knowledge, closely following the network architectures of \citet{Xu/etal/2018}: We first project node features separately into a lower-dimensional space, apply a number of GNN layers $\in \{1, 2, 3, 4, 5\}$ with effective hidden size $\in \{16, 32, 64, 128\}$ and ReLU non-linearity, and perform the final prediction via a fully-connected layer.
All models were implemented using grouped linear projections and evaluated with the number of groups $\in \{1, 8, 16 \}$.

We use the Adam optimizer \citep{Kingma/Ba/2015} with a learning rate of $0.005$ and stop training early with a patience value of $10$.
We apply a fixed dropout rate of $0.5$ before and after GNN layers and add a $\ell_2$ regularization of $0.0005$ to all model parameters.
For our proposed model and GAT, we additionaly tune the number of heads $\in \{ 8, 16 \}$ and set the dropout rate of attention weights to $0.8$.
Hyperparameter configurations of the best performing models with respect to the validation set are reported in Appendix~\ref{sec:hyperparameter_configurations}.

\paragraph{Results.}%
\label{par:results.}

\begin{table}[t]
  \centering
  \caption{
    Results of our DNA approach, in comparison to GCN and GAT with and without Jumping Knowledge.
    Accuracy and standard deviations are computed from 10 random data splits.
  }\label{tab:results}
  \resizebox{\linewidth}{!}{\begin{tabular}{llcccccccc}
    \toprule
      & \mr{2}{\textbf{Model}} & \mr{2}{\textbf{Cora}} & \mr{2}{\textbf{CiteSeer}} & \mr{2}{\textbf{PubMed}} & \textbf{Cora} & \textbf{Coauthor} & \textbf{Coauthor} & \textbf{Amazon} & \textbf{Amazon} \\
      & & & & & \textbf{Full} & \textbf{CS} & \textbf{Physics} & \textbf{Computers} & \textbf{Photo} \\
    \midrule
      \mr{4}{\rb{90}{GCN}} & JK-None   & \gr{83.20 $\pm$ 0.98} & \no{73.87 $\pm$ 0.81} & \gr{86.93 $\pm$ 0.25} & \gr{62.55 $\pm$ 0.60} & \gr{92.90 $\pm$ 0.14} & \gr{95.90 $\pm$ 0.16} & \gr{89.32 $\pm$ 0.20} & \gr{93.11 $\pm$ 0.27} \\
                           & JK-Concat & \gr{83.99 $\pm$ 0.72} & \gr{73.77 $\pm$ 0.89} & \gr{87.52 $\pm$ 0.25} & \no{65.62 $\pm$ 0.49} & \gr{95.44 $\pm$ 0.32} & \gr{96.71 $\pm$ 0.15} & \gr{90.27 $\pm$ 0.28} & \no{94.74 $\pm$ 0.29} \\
                           & JK-Pool   & \no{84.36 $\pm$ 0.62} & \gr{73.86 $\pm$ 0.97} & \no{87.61 $\pm$ 0.27} & \gr{65.14 $\pm$ 0.81} & \bo{95.47 $\pm$ 0.21} & \bo{96.74 $\pm$ 0.17} & \no{90.30 $\pm$ 0.37} & \gr{94.64 $\pm$ 0.24} \\
                           & JK-LSTM   & \gr{80.46 $\pm$ 0.88} & \gr{72.92 $\pm$ 0.69} & \gr{87.38 $\pm$ 0.29} & \gr{55.39 $\pm$ 0.40} & \gr{94.40 $\pm$ 0.28} & \gr{96.55 $\pm$ 0.08} & \gr{90.06 $\pm$ 0.23} & \gr{94.54 $\pm$ 0.30} \\
    \midrule
      \mr{4}{\rb{90}{GAT}} & JK-None   & \bo{86.35 $\pm$ 0.74} & \gr{73.70 $\pm$ 0.53} & \gr{86.76 $\pm$ 0.25} & \gr{65.70 $\pm$ 0.32} & \gr{93.54 $\pm$ 0.17} & \gr{96.21 $\pm$ 0.08} & \gr{88.02 $\pm$ 1.39} & \gr{93.00 $\pm$ 0.42} \\
                           & JK-Concat & \gr{84.70 $\pm$ 0.57} & \no{73.97 $\pm$ 0.46} & \bo{88.73 $\pm$ 0.30} & \no{66.18 $\pm$ 0.47} & \no{95.12 $\pm$ 0.18} & \no{96.66 $\pm$ 0.09} & \no{89.67 $\pm$ 0.59} & \no{94.93 $\pm$ 0.31} \\
                           & JK-Pool   & \gr{83.91 $\pm$ 0.87} & \gr{73.42 $\pm$ 0.71} & \gr{88.44 $\pm$ 0.33} & \gr{61.52 $\pm$ 1.17} & \gr{94.84 $\pm$ 0.16} & \gr{96.62 $\pm$ 0.06} & \gr{89.42 $\pm$ 0.47} & \gr{94.80 $\pm$ 0.24} \\
                           & JK-LSTM   & \gr{78.08 $\pm$ 1.53} & \gr{71.84 $\pm$ 1.20} & \gr{87.85 $\pm$ 0.26} & \gr{55.41 $\pm$ 0.35} & \gr{94.09 $\pm$ 0.23} & \gr{96.45 $\pm$ 0.05} & \gr{87.26 $\pm$ 1.82} & \gr{94.47 $\pm$ 0.33} \\
    \midrule
      \mr{3}{\rb{90}{DNA}} & $g=1$     & \gr{83.88 $\pm$ 0.50} & \gr{73.37 $\pm$ 0.83} & \gr{87.80 $\pm$ 0.25} & \gr{63.72 $\pm$ 0.44} & \gr{94.02 $\pm$ 0.17} & \gr{96.49 $\pm$ 0.10} & \gr{90.52 $\pm$ 0.40} & \gr{94.89 $\pm$ 0.26} \\
                           & $g=8$     & \gr{85.86 $\pm$ 0.45} & \gr{74.19 $\pm$ 0.66} & \no{88.04 $\pm$ 0.17} & \gr{66.50 $\pm$ 0.42} & \gr{94.46 $\pm$ 0.15} & \no{96.58 $\pm$ 0.09} & \bo{90.99 $\pm$ 0.40} & \gr{94.96 $\pm$ 0.24} \\
                           & $g=16$    & \no{86.15 $\pm$ 0.57} & \bo{74.50 $\pm$ 0.62} & \no{88.04 $\pm$ 0.22} & \bo{66.64 $\pm$ 0.47} & \no{94.64 $\pm$ 0.15} & \gr{96.53 $\pm$ 0.10} & \gr{90.81 $\pm$ 0.38} & \bo{95.00 $\pm$ 0.19} \\
    \bottomrule
  \end{tabular}}
\end{table}

Table~\ref{tab:results} shows the average classification accuracy over 10 random data splits and initializations.
Our DNA approach outperforms traditional stacking of GNN layers (JK-None) and even exceeds the performance of Jumping Knowledge in most cases.
Noticeably, the use of grouped linear projections greatly improves attention-based approaches, especially when combined with a large effective hidden size.
We noticed gains in accuracy up to $3$ percentage points when comparing the best results of $g=1$ to $g > 1$, both for GAT and DNA, especially when combined with a large effective hidden size.
Best hyperparameter configurations (\cf~Appendix~\ref{sec:hyperparameter_configurations}) show advantages in using increased feature dimensionalities across all datasets.
For GCN, we found those gains to be negligible.
Similar to JK nets, our approach benefits from an increased amount of stacked layers.

\section{Qualtivate Analysis on Cora}%
\label{sec:qualtivate_analysis_on_cora}

\begin{figure}[b]
  \begin{minipage}[t]{0.52\linewidth}
    \centering
    \caption{
      Influence distributions of different 5-layer GNNs starting at the squared node.
      Due to visibility, we visualize only its 2-hop neighborhood.
    }\label{fig:influence}
    \vspace{-0.2cm}
    \begin{subfigure}[c]{0.44\linewidth}
      \caption{GCN JK-Pool}\label{subfig:gcn_pool}
      \vspace{-0.1cm}
      \includegraphics[width=\linewidth]{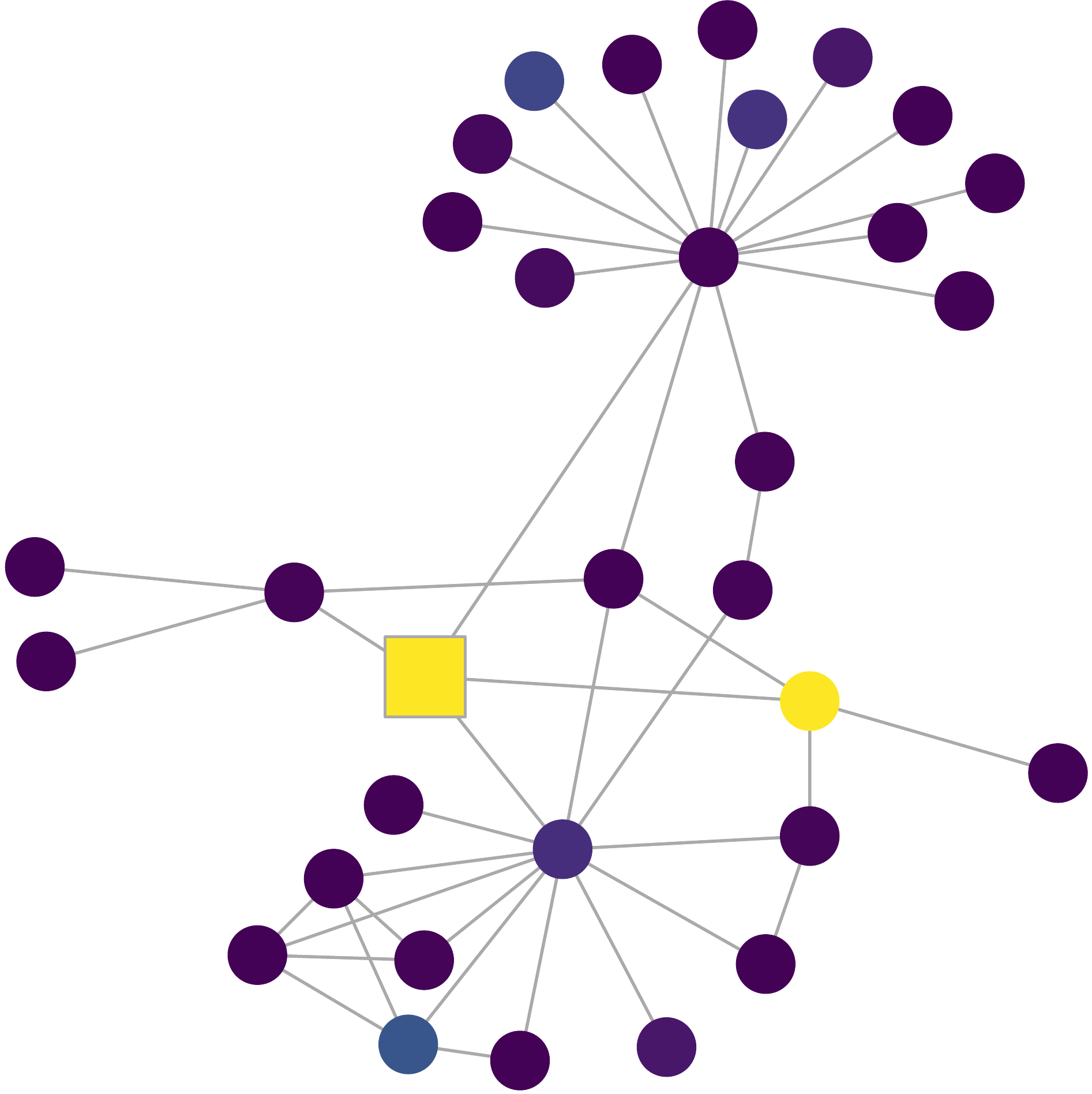}
      \vspace{-0.3cm}
    \end{subfigure}
    \hspace{0.5cm}
    \begin{subfigure}[c]{0.44\linewidth}
      \caption{DNA}\label{subfig:dna}
      \vspace{-0.1cm}
      \includegraphics[width=\linewidth]{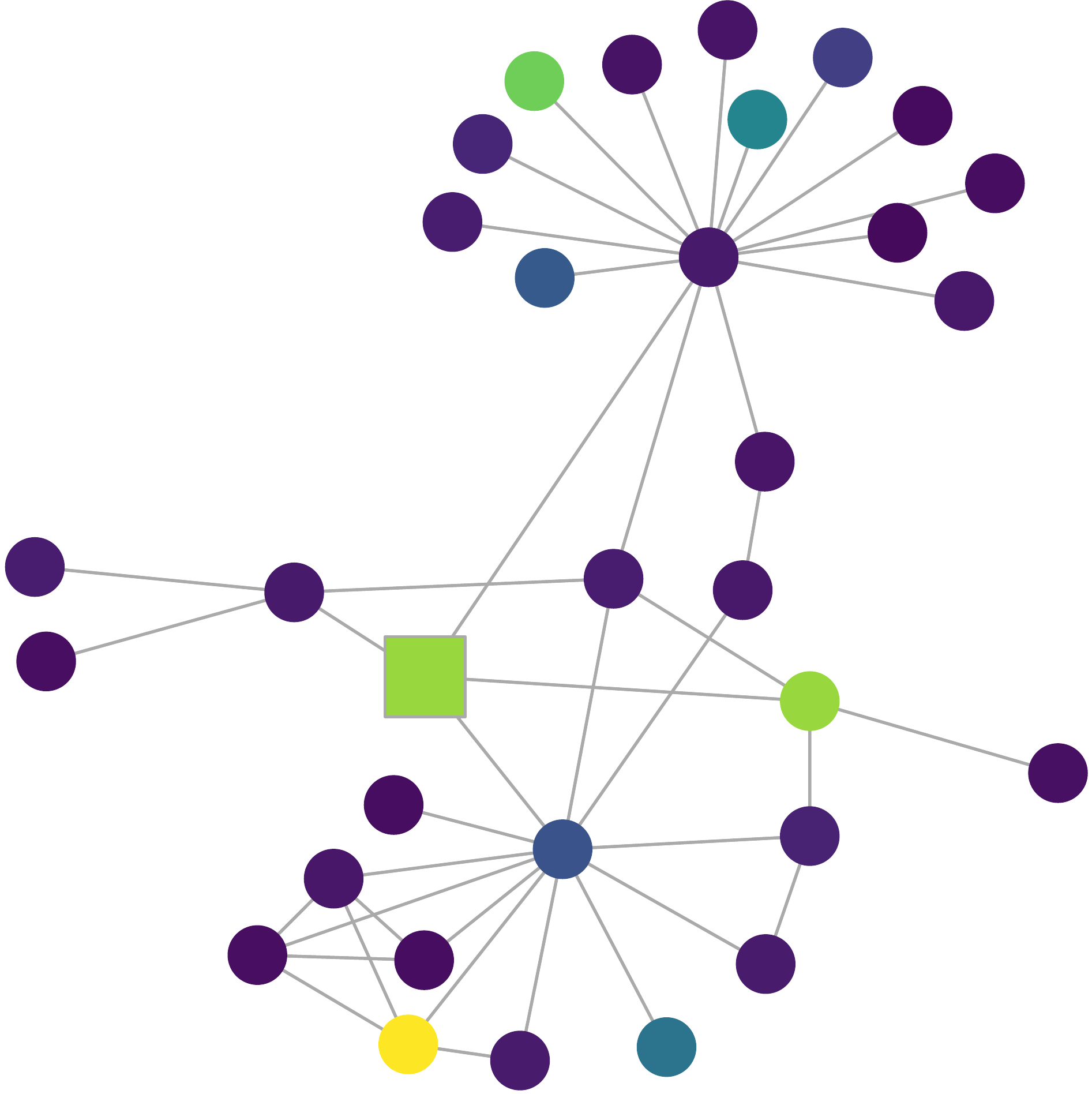}
      \vspace{-0.3cm}
    \end{subfigure}
    \includegraphics[height=0.1cm,width=\linewidth]{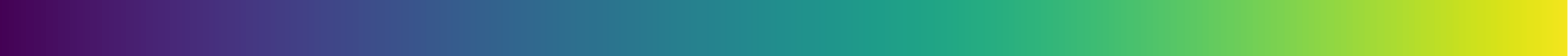}
  \end{minipage}
  \hfill
  \begin{minipage}[t]{0.43\linewidth}
    \centering
    \caption{Final attention weight distribution of a 5-layer DNA-GNN.}\label{fig:boxplot}
    \vspace{-0.2cm}
    \includegraphics[width=0.95\linewidth]{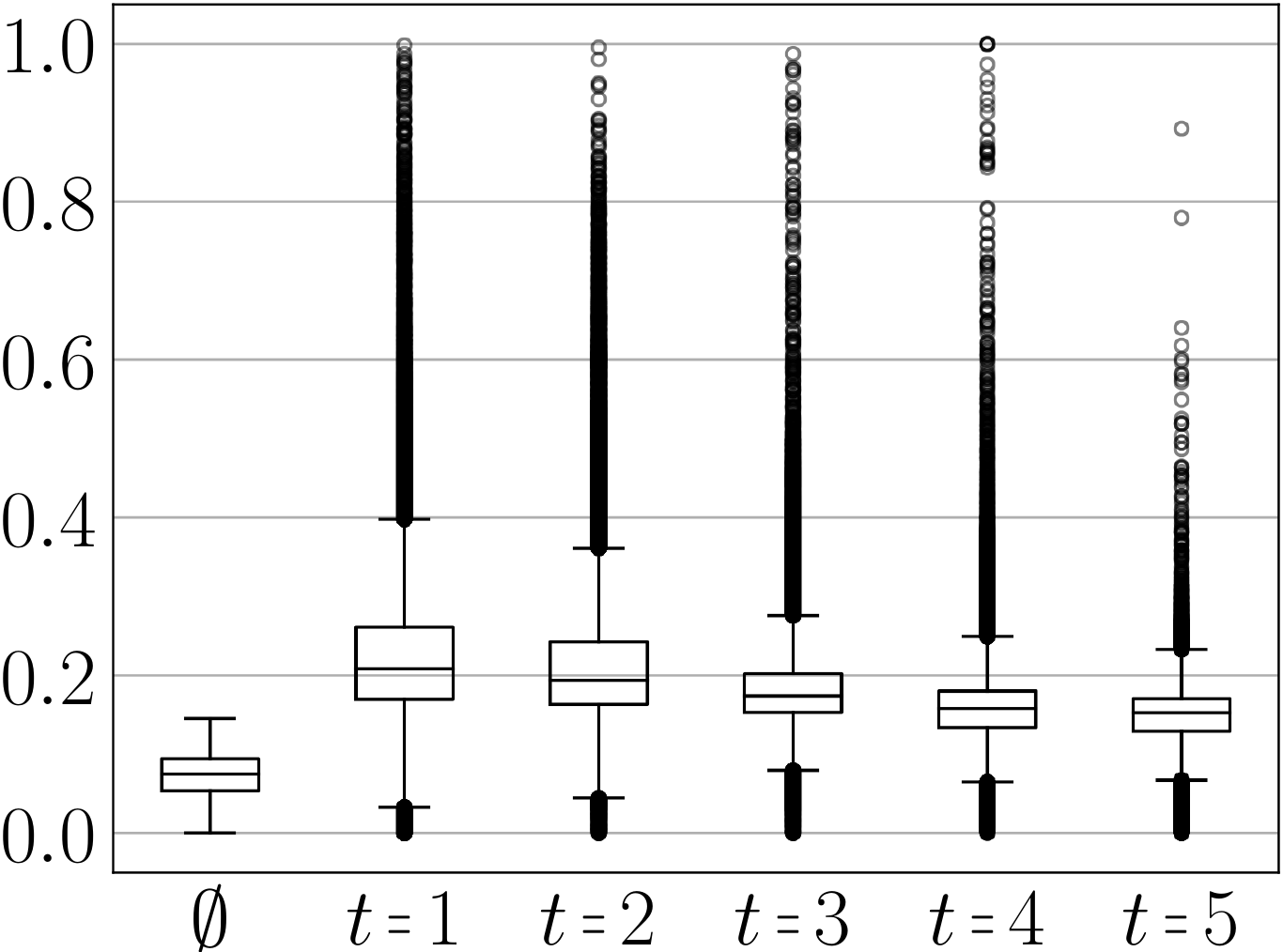}
  \end{minipage}
\end{figure}

We use the (normalized) influence score $I_v(w) = \mathbbm{1}^{\hspace{-2pt} \top} \lmid \frac{\partial \vec{h}_v^{(T)}}{\partial \vec{h}_w^{(0)}} \rmid \mathbbm{1}$ \citep{Xu/etal/2018} to visualize the differences in aggregation starting at a node which is correctly classified by DNA, but is incorrectly classified by GCN JK-Pool (\cf~Figure~\ref{fig:influence}).
While the node embedding of GCN JK-Pool is nearly exclusively influenced by its central node and a node nearby, DNA aggregates \emph{localized} information even from nodes far away.
Figure~\ref{fig:boxplot} signals that aggregations typically attend to earlier representations.
This verifies that nearby information is indeed often sufficient to classify most nodes.
However, there are some nodes that do make heavy usage of information retrieved from latter representations, indicating the merits of a dynamic neighborhood aggregation procedure.

\section{Conclusion}%
\label{sec:conclusion}

We introduced a dynamic neighborhood aggregation (DNA) scheme which computes new embeddings for a node by attending to all previous embeddings of its neighbors.
This dynamic aggregation allows the model to learn to use specific receptive fields and depths for a given task and naturally solves the problems of exponential spread-outs and ``washed out'' representations when naively stacking GNN layers.
In contrast to JK nets, our DNA scheme enables fine-grained node representations in which both local and global information can effectively be combined across different neighborhood branches.
Finally, we showed empirically that grouped operations can be an effective regularizer for attention heads which can additionally enable the usage of larger feature dimensionalities in GNNs.

\subsubsection*{Acknowledgments}

This work has been supported by the \emph{German Research Association (DFG)} within the Collaborative Research Center SFB 876, \emph{Providing Information by Resource-Constrained Analysis}, project A6.
I thank Jan E. Lenssen for proofreading and helpful advice.

\bibliography{iclr2019_conference}
\bibliographystyle{iclr2019_conference}

\newpage

\begin{appendices}

\section{Datasets}%
\label{sec:datasets}

\begin{table}[htb!]
  \centering
  \caption{Dataset statistics of the transductive node-classification experiments.}\label{tab:datasets}
  \begin{tabular}{lrrrrr}
    \toprule
      \textbf{Dataset} & \textbf{Nodes} & \textbf{Edges} & \textbf{Features} & \textbf{Classes} \\
    \midrule
      Cora             & 2,708  & 5,278   & 1,433 & 7   \\
      CiteSeer         & 3,327  & 4,552   & 3,703 & 6   \\
      PubMed           & 19,717 & 44,324  & 500   & 3   \\
      Cora Full        & 19,793 & 63,421  & 8,710 & 70  \\
      Coauthor CS      & 18,333 & 81,894  & 6,805 & 15  \\
      Coauthor Physics & 34,493 & 247,962 & 8,415 & 5   \\
      Amazon Computers & 13,752 & 245,861 & 767   & 10  \\
      Amazon Photo     & 7,650  & 119,081 & 745   & 8   \\
    \bottomrule
  \end{tabular}
\end{table}

Cora, CiteSeer, PubMed and Cora Full \citep{Sen/etal/2008, Bojchevski/Guennemann/2018} are citation network datasets where nodes represent documents, and edges represent (undirected) citation links.
The networks contain bag-of-words feature vectors for each document.

Coauthor CS and Coauthor Physics \citep{Shchur/etal/2018} are co-authorship graphs where nodes are authors which are connected by an edge if they co-authored a paper.
Given paper keywords for each author's paper as node features, the task is to map each author to its most active field of study.

Amazon Computers and Amazon Photo \citep{Shchur/etal/2018} are segments of the Amazon co-purchase graph where nodes represent goods which are linked by an edge if these goods are frequently bought together.
Node feature encode product reviews as bag-of-word feature vectors, and class labels are given by product category.

\section{Hyperparameter Configurations}%
\label{sec:hyperparameter_configurations}

\begin{table}[htb!]
  \centering
  \caption{
    Hyperparameter configuration (number of layers / effective hidden size / number of groups) of the best GCN models with respect to the validation set.
  }\label{tab:gcn_hyperparameters}
  \resizebox{\linewidth}{!}{\begin{tabular}{lcccccccc}
    \toprule
      \mr{2}{\textbf{Model}} & \mr{2}{\textbf{Cora}} & \mr{2}{\textbf{CiteSeer}} & \mr{2}{\textbf{PubMed}} & \textbf{Cora} & \textbf{Coauthor} & \textbf{Coauthor} & \textbf{Amazon} & \textbf{Amazon} \\
      & & & & \textbf{Full} & \textbf{CS} & \textbf{Physics} & \textbf{Computers} & \textbf{Photo} \\
    \midrule
      JK-None   & 1/128/16 & 1/128/8 & 1/16/1  & 1/128/16 & 1/128/16 & 1/32/16 & 1/128/16 & 1/64/16  \\
      JK-Concat & 2/128/8  & 2/64/8  & 2/16/16 & 2/128/8  & 2/128/1  & 3/64/1  & 1/128/1  & 3/128/1  \\
      JK-Pool   & 2/128/1  & 2/128/1 & 2/16/16 & 5/128/16 & 5/128/16 & 5/64/1  & 1/128/8  & 3/128/16 \\
      JK-LSTM   & 1/128/8  & 1/128/1 & 2/16/8  & 1/128/8  & 1/64/1   & 1/64/8  & 1/128/16 & 1/64/1   \\
    \bottomrule
  \end{tabular}}
\end{table}

\begin{table}[htb!]
  \centering
  \caption{
    Hyperparameter configuration (number of layers / effective hidden size / number of groups~/ number of heads) of the best GAT models with respect to the validation set.
  }\label{tab:gat_hyperparameters}
  \resizebox{\linewidth}{!}{\begin{tabular}{lcccccccc}
    \toprule
      \mr{2}{\textbf{Model}} & \mr{2}{\textbf{Cora}} & \mr{2}{\textbf{CiteSeer}} & \mr{2}{\textbf{PubMed}} & \textbf{Cora} & \textbf{Coauthor} & \textbf{Coauthor} & \textbf{Amazon} & \textbf{Amazon} \\
      & & & & \textbf{Full} & \textbf{CS} & \textbf{Physics} & \textbf{Computers} & \textbf{Photo} \\
    \midrule
      JK-None   & 3/128/16/8 & 1/128/16/8 & 1/64/8/8    & 1/128/16/8 & 1/128/8/8  & 1/128/1/8 & 1/128/1/16 & 1/128/8/16 \\
      JK-Concat & 2/128/1/8  & 2/128/1/8  & 5/128/16/8  & 5/128/8/16 & 3/128/1/8  & 2/128/1/8 & 2/128/8/16 & 2/128/8/16 \\
      JK-Pool   & 5/128/1/16 & 4/128/1/16 & 3/128/16/16 & 2/128/1/16 & 2/128/1/16 & 1/128/1/8 & 2/128/1/16 & 2/128/8/16 \\
      JK-LSTM   & 1/128/1/16 & 1/128/1/16 & 2/16/1/8    & 1/128/1/8  & 1/64/1/16  & 1/64/1/8  & 1/64/1/16  & 1/64/1/8   \\
    \bottomrule
  \end{tabular}}
\end{table}

\begin{table}[htb!]
  \centering
  \caption{
    Hyperparameter configuration (number of layers / effective hidden size / number of heads) of the best DNA models with respect to the validation set.
  }\label{tab:dna_hyperparameters}
  \resizebox{\linewidth}{!}{\begin{tabular}{lcccccccc}
    \toprule
      \mr{2}{\textbf{Model}} & \mr{2}{\textbf{Cora}} & \mr{2}{\textbf{CiteSeer}} & \mr{2}{\textbf{PubMed}} & \textbf{Cora} & \textbf{Coauthor} & \textbf{Coauthor} & \textbf{Amazon} & \textbf{Amazon} \\
      & & & & \textbf{Full} & \textbf{CS} & \textbf{Physics} & \textbf{Computers} & \textbf{Photo} \\
    \midrule
      $g=1$  & 1/128/16 & 2/128/8  & 2/16/8  & 2/128/8 & 1/128/16 & 1/32/16  & 2/128/8  & 1/64/8   \\
      $g=8$  & 4/64/8   & 3/128/16 & 2/64/8  & 3/128/8 & 1/64/8   & 1/64/8   & 2/128/16 & 1/128/8  \\
      $g=16$ & 4/128/8  & 4/128/8  & 2/64/16 & 2/128/8 & 1/128/16 & 1/128/16 & 1/128/16 & 1/128/16 \\
    \bottomrule
  \end{tabular}}
\end{table}

\end{appendices}

\end{document}